\pdfoutput=1

\documentclass[11pt]{article}

\usepackage{acl}

\usepackage{times}
\usepackage{latexsym}

\usepackage[T1]{fontenc}

\usepackage[utf8]{inputenc}
\usepackage{caption}
\usepackage{subcaption}
\usepackage{xcolor}

\definecolor{purple}{rgb}{0.5,0,1}
\definecolor{dcyan}{rgb}{0.2,0.6,0.5}
\definecolor{light-gray}{gray}{0.95} 
\definecolor{darkgreen}{RGB}{0,140,0}
\definecolor{darkred}{RGB}{200,0,0}
\definecolor{lightgreen}{RGB}{189,252,192}
\definecolor{lightred}{RGB}{255,205,212}
\definecolor{lightyellow}{RGB}{255,240,160}
\definecolor{lightblue}{RGB}{195,221,255}
\definecolor{lightpurple}{RGB}{232,209,255}

\usepackage{xcolor}

\definecolor{orange}{rgb}{1,0.5,0}
\definecolor{mdgreen}{rgb}{0.05,0.6,0.05}
\definecolor{mdblue}{rgb}{0,0,0.7}
\definecolor{dkblue}{rgb}{0,0,0.5}
\definecolor{dkgray}{rgb}{0.3,0.3,0.3}
\definecolor{slate}{rgb}{0.25,0.25,0.4}
\definecolor{gray}{rgb}{0.5,0.5,0.5}
\definecolor{ltgray}{rgb}{0.7,0.7,0.7}
\definecolor{purple}{rgb}{0.7,0,1.0}
\definecolor{lavender}{rgb}{0.65,0.55,1.0}

\usepackage{arydshln}

\usepackage{paralist}
\usepackage[para]{footmisc}
\usepackage{amsmath}
\usepackage{booktabs}

\newcommand{\exampleParagraph}[2]{\framebox{\parbox{0.90\linewidth}{\paragraph{#1} 
{\footnotesize #2}}}}

\usepackage{microtype}

\usepackage{listings}

\usepackage{graphicx}
\usepackage{stfloats}
\usepackage{tabularx,ragged2e}
\usepackage{multirow}
\usepackage[normalem]{ulem}
\usepackage{xcolor}

\usepackage{color, colortbl}
\definecolor{LightCyan}{rgb}{0.88,1,1}

\definecolor{ecolor}{RGB}{44, 180, 44}
\definecolor{necolor}{RGB}{255,69,0}

\title{Leveraging Data Recasting to Enhance Tabular Reasoning}

\author {
    Aashna Jena\textsuperscript{\rm 1\thanks{Equal Contribution}},
    Vivek Gupta\textsuperscript{\rm {2*}\thanks{Corresponding Author}},
    Manish Shrivastava\textsuperscript{\rm 1},
     Julian Martin Eisenschlos\textsuperscript{\rm 3}
    \\\textsuperscript{\rm 1}LTRC, IIIT Hyderabad;
     \textsuperscript{\rm 2}University of Utah;
     \textsuperscript{\rm 3}Google Research, Zurich;\\
    aashna.jena@research.iiit.ac.in; vgupta@cs.utah.edu;\\ m.shrivastava@iiit.ac.in; eisenjulian@google.com
}

\begin{document}
\maketitle
\begin{abstract}

Creating challenging tabular inference data is essential for learning complex reasoning. Prior work has mostly relied on two data generation strategies. The first is human annotation, which yields linguistically diverse data but is difficult to scale. The second category for creation is synthetic generation, which is scalable and cost effective but lacks inventiveness. In this research, we present a framework for semi-automatically recasting existing tabular data to make use of the benefits of both approaches. We utilize our framework to build tabular NLI instances from five datasets that were initially intended for tasks like table2text creation, tabular Q/A, and semantic parsing. We demonstrate that recasted data could be used as evaluation benchmarks as well as augmentation data to enhance performance on tabular NLI tasks. Furthermore, we investigate the effectiveness of models trained on recasted data in the zero-shot scenario, and analyse trends in performance across different recasted datasets types.


\end{abstract}

\section{Introduction}

Given a premise, Natural Language Inference (NLI) is the task of classifying a hypothesis as entailed (true), refuted (false) or neutral (cannot be determined from given premise). Several large scale datasets such as SNLI \cite{snli}, MultiNLI \cite{mnli}, and SQuAD \cite{squad} explore NLI with unstructured text as the premise. 

While textual inference is commonly researched, structured data (e.g. tables, knowledge-graphs and databases) enables the addition of more complex types of reasoning, such as ranking, counting, and aggregation. Creating challenging large scale supervision data is vital for research in tabular reasoning. In recent years, many initiatives to include tabular semi-structured data as the premise have been introduced e.g. tabular inference (TNLI) datasets such as TabFact \cite{2019TabFactA}, InfoTabS \cite{Gupta:20} and shared tasks like SemEval 2021 Task 9 \cite{Wang2021SemEval2021T9} and FEVEROUS \cite{aly-etal-2021-fact}. Tabular data differs from unstructured text in the way that it can capture information and relationships in a succinct manner through underlying structure \citep{Gupta:20}.

\begin{table}
\begin{center}
\begin{subtable}{0.42\textwidth}
\small
\hspace{10mm}
\begin{tabular}{|c|c|c|}
\multicolumn{3}{c}{\textbf{Example Table}} \\
\hline
Party&Votes(thou)& Seats\\
\hline
Party A&650&120 \\
\cellcolor{green!25}Party B&570&\cellcolor{green!25}89 \\
  Party C&TBA&89 \\
  Total  &1235&298 \\
\hline
\end{tabular}
\end{subtable}
\end{center}
\begin{subtable}{0.48\textwidth}
\begin{center}
\small
\begin{tabular}{|c|}
\hline
\textit{\textbf{Q}: How many seats did Party B win? ; \textbf{A}: 89} \\
\bigg \downarrow \textbf{recast} \bigg \downarrow \\
\textit{\textcolor{darkgreen}{Entailment}: Party B won 89 seats.}\\
\textit{\textcolor{red}{Contradiction}: Party B won 120 seats.} \\
\hline
\end{tabular}
\end{center}
\end{subtable}
\caption{\label{table:intro}Example of Tabular Data Recasting}
\vspace{-3mm}
\end{table}

Despite fluent, diverse, and creative, these human-annotated datasets are limited in scale owing to the costly and time-consuming nature of annotations. Furthermore, \citet{gururangan-etal-2018-annotation} and \citet{geva-annotator} show that many human-annotated datasets for NLI contain annotation biases or artifacts. This allows NLI models to learn spurious patterns \citep{niven-kao-2019-probing}, which enables models to predict the right label for the wrong reasons, sometimes even with noisy, incorrect or incomplete input \cite{poliak2018hypothesis}. Recently, \citet{table-probing} revealed that tabular inference datasets also suffer from comparable challenges. Furthermore, \citet{geva-annotator,parmar2022don} show that annotators introduce their own bias during annotation. For example, \citet{table-probing} demonstrates that annotators only generate hypothesis sentences from keys having numerical values, implying that some keys are either over or underutilized.

On the other hand, automatic grammar-based strategies , despite their scalability, lack linguistic diversity (both structural and lexical) and necessary inference complexity. These approaches create examples with only naive reasoning. Recently, \cite{geva, julian}, create  context-free-grammar and templates to generate augmentation data for Tabular NLI. Additionally, the use of a large language generation model ~\cite[e.g.,][]{radford2018improving,lewis-etal-2020-bart,raffel2020exploring} for data generation has been proposed as well ~\cite[e.g.,][]{radford2018improving, ouyang2022training,naturalinstructionsv1}. However, such generation systems lack factuality, leading in hallucinations, inadequate fact coverage, and token repetition problem.

\paragraph{\textit{Can we generate challenging supervision data that is as scalable as synthetic data and yet contains human-like fluency and linguistic diversity?}} In this work, we attempt to answer the above question through the lens of data recasting. Data recasting refers to transforming data intended for one task into data intended for another distinct task. Although data recasting has been around for a long time, for example, QA2D \citep{Demszky2018TransformingQA} and SciTail \cite{scitail} effectively recast question answering data for inference (NLI), no earlier study has applied it to semi-structured data. 

Therefore, we propose a semi-automatic framework for tabular data recasting.  Using our framework, we generate large-scale tabular NLI data by recasting existing datasets intended for non NLI tasks such as Table2Text generation (T2TG), Tabular Question Answering (TQA), and Semantic Parsing on Tables (SPT). This recasting strategy is a middle road technique that allows us to benefit from both synthetic and human-annotated data generation approaches. It allows us to minimise annotation time and expense while maintaining linguistic variance and creativity via human involvement from the original source dataset. \autoref{table:intro} shows an example of tabular data recasting. Note that while the steps we describe for recasting in this work are automatic, we call the overall end-to-end framework \textbf{semi}-automatic to include the manual effort gone into creation of the source datasets.

Our recasted data can be used for both evaluation and augmentation purposes for tabular inference tasks. Models pre-trained on our data show an improvement of 17\% from the TabFact baseline \cite{2019TabFactA} and 1.1\% from \citet{julian}, a synthetic data augmentation baseline. Additionally, we report a zero-shot accuracy of 71.1\% on the TabFact validation set, which is 5\% percent higher than the supervised baseline accuracy reported by \citet{2019TabFactA}.

Our main contributions are the following: 

\begin{enumerate}
\itemsep0em 
\item We propose a semi-automatic framework to generate tabular NLI data from other non-NLI tasks such as T2TG, TQA, and SPT. 

\item We build five large-scale, diversified, human-alike, and complex tabular NLI datasets sourced from datasets as shown in \autoref{tab:dataRecastSource}.

\item We present the usage of recasted data as TNLI model evaluation benchmarks. We demonstrate an improvement in zero-shot transfer performance on TabFact using recasted data.  We demonstrate the efficacy of our generated data for data augmentation on TabFact.

\end{enumerate}

The dataset and associated scripts are available at \url{https://recasting-to-nli.github.io}

\section{Why Table Recasting?}

Tables are structured data forms. Table cells define a clear boundary for a standalone independent piece of information. These defined table entries facilitate the task of drawing alignments between relevant table cells and given hypotheses. If both the premise and hypothesis were plain text, any n-gram in the premise could be aligned to any n-gram in the hypothesis. Finding alignments between the premise and hypothesis is crucial in the recasting process. This is because, in order to modify statements, we must understand which entities influence their truth value.

Moreover, in tables, entries of the same type (same part of speech type, named entity type, domain etc) are clubbed under a common column header. This allows us to easily identify a group of candidates which are interchangeable in a sentence without disrupting its coherence. Frequently, the column header also indicates the data type (e.g. Name, Organization, Year etc). This is incredibly beneficial when modifying source data by substituting entities.
\begin{table*}[t]
\small
\begin{center}
\parbox{.40\linewidth}{
\centering
\begin{tabular}{|c|c|c|}
\multicolumn{3}{c}{\textbf{Original Table (OG)}} \\
\hline
Party&Votes(thou)& Seats\\
\hline
\cellcolor{green!25}Party A&650&\cellcolor{green!25}120 \\
  Party B  &570&89 \\
  Party C  &final count TBA&89 \\
  Total  &1235&\cellcolor{green!25}298 \\
\hline
\end{tabular}
}
\parbox{.46\linewidth}{
\centering
\begin{tabular}{|c|c|c|}
\multicolumn{3}{c}{\textbf{Counterfactual Table (CF - after cells swaps)}} \\
\hline
Party&Votes(thou)& Seats\\
\hline
\cellcolor{blue!25}\sout{Party A} \textbf{Party B} &650&\cellcolor{blue!25}120 \\
\sout{Party B} \textbf{Party A}&570&89 \\
Party C&final count TBA&89 \\
Total&1235&\cellcolor{blue!25}298 \\
\hline
\end{tabular}
}   
\end{center}

\begin{center}
\vspace{-1.0em}
\parbox{0.9\linewidth}{
\centering
\begin{tabular}{|rll|}
\multicolumn{2}{c}{} \\
\hline
\textit{Base Entailment\textsubscript{OG}}& \colorbox{green!25}{Party A} won \colorbox{green!25}{120} out of \colorbox{green!25}{298} seats.&\colorbox{green!25}{Party A} won the \colorbox{green!25}{most} seats.\\
\textit{New Entailment\textsubscript{OG}}&\textcolor{blue}{Party B} won \textcolor{blue}{89} out of \textcolor{blue}{298} seats.&\textcolor{blue}{Party B} won the \textcolor{blue}{second most} seats.\\
\textit{Paraphrase\textsubscript{OG}}&Out of a total of \textcolor{blue}{298} available seats,&\textcolor{blue}{Party B} secured the \textcolor{blue}{second largest}\\
&\textcolor{blue}{Party B} won \textcolor{blue}{89}.& number of seats.\\
\textit{Contradiction\textsubscript{OG}}&\sout{Party A} \textcolor{red}{Party B} won 120 out of 298&\sout{Party A} \textcolor{red}{Party B} won the most seats.\\
 &seats.&Party A won the \sout{most} \textcolor{red}{least} seats.\\
\multicolumn{3}{|c|}{} \\
\multicolumn{3}{|p{13cm}|}{\textit{We swap \textbf{Party A} and \textbf{Party B} to create a counterfactual table. The contradictions mentioned above 
become the new base annotations (Annotation\textsubscript{CT})}} \\
\multicolumn{3}{|c|}{} \\

\textit{Base Entailment\textsubscript{CF}}& \colorbox{blue!25}{Party B} won \colorbox{blue!25}{120} out of \colorbox{blue!25}{298} seats.&\colorbox{blue!25}{Party B} won the \colorbox{blue!25}{most} seats.\\

\textit{New Entailment\textsubscript{CF}}&\textcolor{blue}{Party A} won \textcolor{blue}{89} out of \textcolor{blue}{298} seats.& \textcolor{blue}{Party A} won the \textcolor{blue}{second most} seats.\\

\textit{Paraphrase\textsubscript{CF}}&\textcolor{blue}{89} of the \textcolor{blue}{298} available seats were& \textcolor{blue}{Party A} won \textcolor{blue}{next to the maximum}\\

&secured by \textcolor{blue}{Party A}& number of seats.\\

\textit{Contradiction\textsubscript{CF}}&\sout{Party B} \textcolor{red}{Party A} won 120 out of 298&\sout{Party B} \textcolor{red}{Party A} won the most seats.\\
 &seats.&Party B won the \sout{most} \textcolor{red}{least} seats.\\
\hline
\end{tabular}
}
\end{center}
\vspace{-1.0em}
\caption{Pipeline for generating recasted NLI data. We first create entailments and contradictions from the given base annotation. We then create a counterfactual table taking a contradiction to be the new base annotation. subscript\textsubscript{OG} represents the ``Original'' table and subscript\textsubscript{CF} represents the ``Counterfactual'' table. Note that Base Entailment\textsubscript{OG} contradicts Table\textsubscript{CF} and Base Entailment\textsubscript{CF} contradicts Table\textsubscript{OG}. This pair will always exhibit this property, but there can be statements which entail (or contradict) both OG and CF tables.}
\label{table:table1}
\end{table*}

\section{Tabular Recasting Framework}
\label{sec:frameworkSection}

In this section, we describe a general semi-automatic framework to recast tabular data for the task of Table NLI. By recasting, we imply converting data meant for one task into a format that satisfies the requirements of a different task.

\subsection{Prerequisites}
A Table NLI data instance consists of (a) a \textbf{table}, (b) some \textbf{entailments}, i.e. true claims based on the table, and (c) some \textbf{contradictions}, i.e. false statements based on the table. To be able to generate these, a table is the initial prerequisite. Since we utilise tabular data as our source, this need is met. 

In addition to the table, we require at least one reference statement that validates the table. We utilize the structure of this reference statement (henceforth referred to as the \textbf{Base Entailment}) to generate further entailments and contradictions. Once we have the Base Entailment, contradictions can be formed fairly easily. Falsifying any part of the Base Entailment that is linked to the table creates contradictions. 

To create an entailment, however, every portion of the perturbed statement must hold true for the entire statement to constitute an entailment. This means that all entities originating from the table (henceforth referred to as \textbf{relevant entities}) must be found in the Base Entailment. Then and only then can we know with certainty how perturbations affect the truth value of a given assertion.

Alignments between a table and a Base Entailment are not always apparent, as demonstrated in \autoref{table:table1}. In the example \textit{"Party A won the \textbf{most} seats}, the alignment between \textit{most} and the greatest number of seats must be determined. Although we can employ automatic matching techniques between the Base Entailment and the table to extract relevant entities, we cannot be certain of detecting \textbf{all} of them unless they are explicitly provided. Therefore, we must be able to extract the following from source datasets: (a) a \textbf{table} i.e. the Premise, (b) a \textbf{reference statement} i.e. the Base Entailment, (c) \textbf{relevant entities} and (d) their \textbf{alignments} with the reference statement.

Once the prerequisites are met, new NLI instances can be formed by perturbing existing data in two ways: (a) by perturbing the hypothesis and (b) by perturbing the table, i.e. the premise.

\subsection{Perturbing the Hypothesis}

We modify the hypothesis, i.e. the Base Entailment, by substituting relevant entities with other potential candidates. We presume the tables are vertically aligned, which means that the top row contains headers and each column contains entities of the same kind. A \textit{potential candidate} for a relevant entity coming from table cell \(C\textsubscript{XY}\) having coordinates \([\text{row}~X, \text{column}~Y]\) can be any other non-null entity from the same column \(Y\). 

\noindent \paragraph{Creating Entailments (E)} To create entailments, we replace \textit{all} the relevant entities in the given Base Entailment with potential candidates. Two or more relevant entities coming from table cells in the same row, say \(C\textsubscript{XA}, C\textsubscript{XB}\), must be substituted with potential candidates from column \(A\) and \(B\) respectively, such that their row coordinate is equivalent i.e. \(C\textsubscript{XA}, C\textsubscript{XB} \rightarrow C\textsubscript{ZA}, C\textsubscript{ZB} \mid Z \neq X\) (refer \autoref{table:table1}). Entities originating from “aggregate rows” (such as the \textit{Total} row in \autoref{table:table1}) or "headers" must be left intact.

\noindent \paragraph{Creating Contradictions (C)} To create contradictions, we substitute \textbf{one or more} relevant entities from the Base Entailment with alternative candidates. We note that the ensuing statement may be an entailment by accident. In \autoref{table:table1}, consider the Base Entailment - \textit{“\textbf{Party B} won 89 seats”}. Suppose we replace one key entity (Party B) with a potential candidate to arrive at “\textit{\textbf{\sout{Party B} Party C}} won 89 seats”. The resultant statement remains an entailment. To prevent this from occurring, the non-replaced entities are compared. Assume \(C\textsubscript{XA}, C\textsubscript{XB}\) represent the relevant entities in the Base Entailment. If we replace \(C\textsubscript{XA} \rightarrow C\textsubscript{ZA}\) then we must guarantee that \(C\textsubscript{XB} \neq C\textsubscript{ZB}\) to avoid unintentional entailments.

We also generate contradictions by substituting antonyms for words in the Base Entailment. This is particularly helpful for scenarios involving superlatives and comparatives (refer \autoref{table:table1}). We use NLTK Wordnet to find antonyms. We enforce equality of POS tags for word-antonym pairs. We create a dictionary of such pairs across datasets, count their frequency and broadly filter (manually) the word-antonym pairs for keeping the frequent and sensible ones. Majority of these are comparative/superlative word-pairs (higher-lower, most-least, best-worst).

\subsection{Perturbing the Table (Premise)}

In this subsection, instead of modifying the Base Entailment, we swap two or more table cells to modify the premise, i.e. the tables. Similar to \citet{kaushik2020learning} and \citet{gardner2020evaluating}, we build example pairs with minimal differences but opposing inference labels in order to improve model generalisation. These modified tables no longer reflect the actual world. Hence, we refer to them as \textbf{Counterfactual}. The addition of counterfactual data increases the model's robustness by preventing it from learning spurious correlations between label and hypothesis/premise. Minimally varying counterfactual data also ensures that the model is not biased and preferably grounds on primary evidence, as opposed to depending blindly on its pre-trained knowledge. Similar findings were made by \citet{muller2021tapas} for TabFact.

\noindent \paragraph{Creating Counterfactual Tables (CF)} We consider a contradiction C1 formed by replacing the relevant cell \(C\textsubscript{XA} \rightarrow C\textsubscript{ZA}\) in the original table (as described in \autoref{table:table1}). To create a counterfactual table, we swap cells \(C\textsubscript{XA} \leftrightarrow C\textsubscript{ZA}\) such that C1 becomes an entailment to the modified table, and the original Base Entailment becomes a contradiction to it. Based on this, we generate further hypotheses, as illustrated in \autoref{table:table1}.

\noindent \paragraph{Hypothesis Paraphrasing (HP)} 
\citet{Dagan:13} demonstrates that data paraphrasing increases lexical and structural diversity, thus boosting model performance on unstructured NLI. In accordance with \citet{Dagan:13}, we paraphrase our data because the hypotheses derived from Base Entailments have similar structures. For producing paraphrases, we employ the publicly available T5 Model \cite{raffel2020exploring} trained on the Google PAWS dataset \cite{paws2019naacl}. We produce the top five paraphrases and then select at random from among them.

\section{Addressing Tabular Recasting Constraints}
\label{sec:challenges}

Dataset specific implementations pose several challenges. We address them in the following ways:

\paragraph{Table Orientation.} As stated in Section \ref{sec:frameworkSection}, we conduct all experiments assuming the tables are vertically aligned. We observe several horizontally aligned tables (with the first column containing headers) in source datasets. As a preliminary processing step, we employ heuristics to automatically recognise such tables and flip them. For example, we check for frequently occurring header names in the first column or consistency in data types (numeric, alpha, etc.) across rows rather than columns.

\paragraph{Partial Matching.} We observe that some datasets provide relevant cells, but do not provide their alignments with the Base Entailment. We attempt to match every relevant cell with n-grams in the Base Entailment. Of particular interest is the sample row shown in \autoref{table:table2} that contains names, numbers, locations and dates that are not exact, but partial matches to n-grams in the Base Entailment. We handle such cases of partial matching.

\begin{table} [!h]
\small
\centering
\begin{tabular}{lr}
\hline
\multicolumn{2}{c}{\textbf{US presidential inaugurations} (A table row)} \\
\hline
\textit{President \#} & 44\\
\textit{Name} & Barack Obama\\
\textit{Inauguration Date} & January 20, 2009\\
\textit{Location} & West Front, US Capitol\\
\hline
\multicolumn{2}{p{7cm}}{\textit{Base Entailment :}} \\
\multicolumn{2}{p{7cm}}{\textbf{Obama's} inauguration as the \textbf{forty fourth} president took place at the \textbf{United States Capitol} in \textbf{2009}.} \\
\hline
\end{tabular}
\caption{\label{table:table2}An example of cases requiring partial matching.}
\end{table}

\paragraph{Irreplaceable Entities.} We observe that \textbf{not all} relevant entities are replaceable by potential candidates. \autoref{table:table1} presents an example of a table with a \textbf{Total} row. Relevant entity \colorbox{green!25}{298} cannot be replaced while creating \textit{New Entailment\textsubscript{OG}} because it is an aggregate entity that whose substitution will disrupt the truth value of the statement.

Similar observation is made while swapping table cells to create counterfactual tables. Suppose we swap the aggregate cell \colorbox{green!25}{298} with \colorbox{green!25}{120}. The resultant table would be logically flawed since the "Seats" column won't add up to its Total. To prevent this, aggregate rows and header cells are marked as non-replaceable entities.

\section{Dataset Recasting}
\label{sec:autotnlidataset}

Using the framework outlined in Section \ref{sec:frameworkSection}, we recast the five datasets listed in \autoref{tab:dataRecastSource}. All datasets utilise open-domain Wikipedia tables, comparable to TabFact. In addition, these datasets and TabFact share reasoning kinds such as counting, minimum/maximum, ranking, superlatives, comparatives, and uniqueness, among others.  \autoref{table:dataset-stats} summarises the statistics of recasted datasets. \footnote{Some of these datasets contain examples that are shared, but because the derivation procedure for NLI data is unique for each task type, generated statements are also different and regarded as individual instances.}

\begin{table}[t]
\small
\centering
\begin{tabular}{ll}
\hline
\textbf{Source Dataset} & \textbf{Task} 
\\
\hline

WikiTableQuestions \citep{pasupat2015compositional} & TQA\\  
FeTaQA \citep{nan-etal-2022-fetaqa} & TQA \\
Squall \citep{Shi2020OnTP} & SPT \\ 
WikiSQL \citep{Zhong2017Seq2SQLGS} & SPT\\
ToTTo \citep{parikh2020totto} & T2TG\\
\hline
\end{tabular}
\vspace{-1.0em}
\caption{Source datasets used for creating tabular NLI data}
\label{tab:dataRecastSource}
\end{table}

\begin{table} [t]
\small
\centering
\begin{tabular}{lrrr}
\hline
\textbf{Dataset} & 
\textbf{Entail} &
\textbf{Contradict} &
\textbf{Total}\\
\hline
QA-TNLI& 32k & 77k & 109k\\
WikiSQL-TNLI& 300k & 385k & 685k\\
Squall-TNLI& 105k & 93k & 198k\\
ToTTo-TNLI& 493k & 357k & 850k \\
\hline
\end{tabular}
\caption{\label{table:dataset-stats} Statistics for various recasted datasets. QA-TNLI combines recasted data from both FeTaQA and WikiTableQuestions. Test splits are created by randomly sampling 10\% samples from each dataset.}
\end{table}

\subsection{Table2Text Generation to Table NLI}
Given a table and a set of highlighted cells, the Table2Text generation task is to create a description derived from the highlighted cells. We presume this description to be the \textit{Base Entailment} given that it is true based on the table. The highlighted cells become the \textit{relevant entities}. An example is shown in \autoref{table:table1}, where \textit{Base Entailment \textsubscript{OG}} is a description generated from OG Table's \colorbox{green!25}{highlighted cells}. 

\paragraph{Recasting ToTTo \citep{parikh2020totto}.} 
ToTTo comprises of over 120k training examples derived from Wikipedia tables. Annotators edit freely written Wikipedia text to produce table descriptions. Annotators also mark relevant cells, but not their alignments with the Base Entailment. To link relevant cells with tokens in the Base Entailment, we apply partial matching techniques. If all relevant cells are successfully matched, we proceed to build new entailments. In either scenario, contradictions are generated using any relevant cells for which alignments can be found. \autoref{table:table2} illustrates this with an example.

\subsection{Table Question Answering To Table NLI}
\label{subsection:qa}
Given a table and a question, the Table Question Answering task is to generate a long form (sentence) or short form (one word/phrase) answer to the question. A long form answer is a \textit{Base Entailment} in itself. \autoref{table:intro} depicts an example of recasting QA data.

\paragraph{Recasting WikiTableQuestions \citep{pasupat2015compositional}.} WikiTableQuestions provides 22k questions over Wikipedia tables, with short-form answers. We use a T5 based pre-trained model developed by \citet{chen-etal-2021-nli-models} to convert \(\{Question, Answer\} \rightarrow Statement\) (refer \autoref{table:intro}). We presume this to be our Base Entailment. Unless it is an aggregate value, the short-form answer is likely to be an entity from the table. We search for matches between the answer and table cells as well as n-grams in the question. We create contradictions from any relevant entities found.

\paragraph{Recasting FeTaQA \citep{nan-etal-2022-fetaqa}.}
FeTaQA provides 10k question-answer pairs on Wikipedia tables. These long form answers are Base Entailments in themselves. Since supporting cell information is provided as well, we create both entailments and contradictions wherever possible.

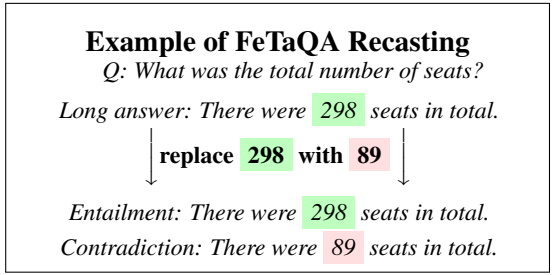
\begin{center}
\noindent \exampleParagraph{Example of FeTaQA Recasting\\}
{
\small
{\centering
\textit{ \smallskip Q: What was the total number of seats?} \\
\textit{Long answer: There were \colorbox{green!25}{298} seats in total.} \\
\bigg \downarrow \textbf{replace \colorbox{green!25}{298} with \colorbox{pink!50}{89}} \bigg \downarrow \\ \smallskip
\textit{Entailment: There were \colorbox{green!25}{298} seats in total.} \\
\textit{Contradiction: There were \colorbox{pink!50}{89} seats in total.} \\
\par}}
\end{center}

\subsection{Semantic Parsing to Table NLI}
Given a table and a question, the Semantic Parsing task is to generate the underlying logical/SQL form of the question. Since the datasets provide a logical query form, we execute this query to obtain a ``short-form answer''. We combine the question and short form answer as mentioned in \ref{subsection:qa} to get the \textit{Base Entailment}.
SQL queries are parsable, allowing for easy identification of column names and cell values. Owing to the fact that the reasoning depends on these entities, we infer these are the \textit{relevant entities}. 

\paragraph{Recasting WikiSQL \citep{Zhong2017Seq2SQLGS}.} WikiSQL provides 67.7k annotated [SQL query, textual question] pairs on Wikipedia tables. To augment this data, we parallelly replace values in an SQL query and its corresponding question. We execute the new query, and combine the answer with the perturbed question to create a new entailment.

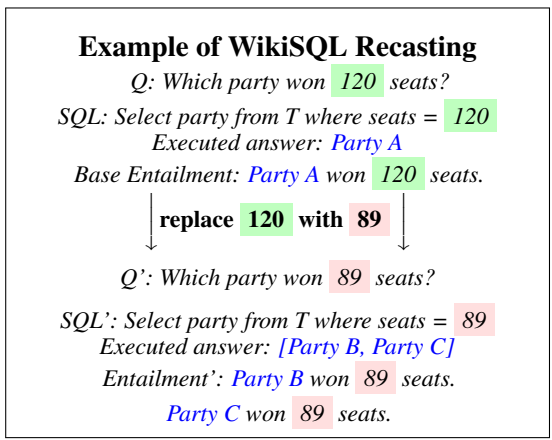
\begin{center}
\noindent \exampleParagraph{Example of WikiSQL Recasting\\}{ \small
{\centering
\textit{Q: Which party won \colorbox{green!25}{120} seats? \linebreak
SQL: Select party from T where seats = \colorbox{green!25}{120} \linebreak
Executed answer: \textcolor{blue}{Party A} \linebreak
Base Entailment: \textcolor{blue}{Party A} won \colorbox{green!25}{120} seats.} \linebreak \smallskip
\bigg \downarrow \textbf{replace \colorbox{green!25}{120} with \colorbox{pink!50}{89}} \bigg \downarrow \linebreak \smallskip
\textit{Q': Which party won \colorbox{pink!50}{89} seats? \linebreak
SQL': Select party from T where seats = \colorbox{pink!50}{89} \linebreak
Executed answer: \textcolor{blue}{[Party B, Party C]} \linebreak
Entailment': \textcolor{blue}{Party B} won \colorbox{pink!50}{89} seats. \linebreak
\textcolor{blue}{Party C} won \colorbox{pink!50}{89} seats.
}\par}}
\end{center}

Note that when executing a query, the answer can be a single entity or a list of multiple entities. If we have a list of entities satisfying the query, any of these entities can be used to create entailments, while none of these entities should be used to create contradictions (we find other potential candidates from the answer column). Consider the OG table
given in \autoref{table:table1}.

\paragraph{Recasting Squall.} Squall provides 11k [sql query, textual question] pairs with table metadata. We augment it similar to WikiSQL. Furthermore, table metadata enables us to identify column kinds and, in some circumstances, reduce SQL queries and questions to skeletons. These skeletons may subsequently be used to generate hypotheses on additional tables that meet the column type specifications of the skeleton in question.  Consider the example from \autoref{table:table1} with columns Party (text) and Seats (numeric).

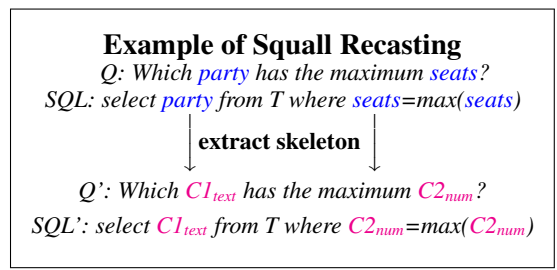
\begin{center}
\small
\noindent \exampleParagraph{Example of Squall Recasting\\}{\small
{\centering
\textit{Q: Which \textcolor{blue}{party} has the maximum \textcolor{blue}{seats}? \linebreak
SQL: select \textcolor{blue}{party} from T where \textcolor{blue}{seats}=max(\textcolor{blue}{seats})} \linebreak
\smallskip
\bigg \downarrow \textbf{extract skeleton} \bigg \downarrow \linebreak \smallskip
\textit{Q': Which \textcolor{magenta}{C1\textsubscript{text}} has the maximum \textcolor{magenta}{C2\textsubscript{num}}? \linebreak
SQL': select \textcolor{magenta}{C1\textsubscript{text}} from T where \textcolor{magenta}{C2\textsubscript{num}}=max(\textcolor{magenta}{C2\textsubscript{num}})\linebreak
}\par}}
\end{center}

This can now be used on another table, suppose one about countries and their populations to ask "Which \textcolor{magenta}{country} has the maximum \textcolor{magenta}{population}?".

\subsection{Human Evaluation}
We asked five annotators to annotate fifty samples from each dataset on two fronts:
\begin{itemize}
    \item Inference label: Label each sample as entail, refute or neutral. Neutral samples can either be those which can't be derived from the table, or which don't make sense.
    \item Coherence score: Score each sample on a scale of 1 to 3 based on its semantic coherence and grammatical correctness, 1 being incoherent and 3 being coherent with minor or no grammatical issues. A score of 2 is given to statements whose meaning can be understood, but the structure or grammar is incorrect in more than one place.
\end{itemize}
We compare our generated label with the majority annotated inference label, and if no majority was reached, we consider the sample inconclusive. For Coherence score, we calculate the average of the five annotators. 

\textbf{Analysis.} Results are summarized in \autoref{tab:human-verification}. We observe high label match scores for our datasets, with QA-TNLI at 90\%, Squall-TNLI at 87\% and WikiSQL-TNLI at 84\%. ToTTo-TNLI is slightly behind at 78\%, which is largely due to samples marked as ``neutral'' or samples where no majority was reached. We also observe a consistently above average coherence score, largely between 2.5 and 3. This implies that most of our data is logical, coherent, and grammatical. Since the sources of our data are human-written (Wikipedia text/human annotations), we expect our generated sentences to be fluent and semantically correct. 

\begin{table}[t]
\small
\centering
\begin{tabular}{lcc}
\hline
\textbf{Dataset} & \textbf{\% Label Match} & \textbf{Coherence Score}
\\
\hline
QA-TNLI & 90\% & 2.68\\
Squall-TNLI & 87\% & 2.54\\ 
WikiSQL-TNLI & 84\% & 2.55\\
ToTTo-TNLI & 78\% & 2.46\\
\hline
\end{tabular}
\vspace{-1.0em}
\caption{\small Results for human evaluation of our generated data. Please note that the verification labels are considered to be matched only if annotators have reached a majority \textbf{and} it matches our generated label.}
\label{tab:human-verification}
\end{table}



\section{Experiments and Analysis}

In this section, we examine the relevance of our recast data across various settings. Overall, we aim to answer the following research questions:
\begin{enumerate}
    \item \textbf{RQ1}: How challenging is recast data as a TNLI benchmark?
    \item \textbf{RQ2}: How effective are models trained on recasted data in a zero shot setting?
    \item \textbf{RQ3}: How beneficial is recasted data for TNLI data augmentation?
\end{enumerate}

\subsection{Experimental Setup}
\label{subsection:model}
In all experiments, we follow the pre-training pipeline similar to \citet{julian}. We start with TAPAS \cite{herzig-etal-2020-tapas}, a table-based BERT model, and intermediately pre-train it on our recasted data. We then fine-tune the model on the downstream tabular NLI task.

\paragraph{Dataset.} We use TabFact \cite{2019TabFactA}, a benchmark Table NLI dataset, as the end task to report results. TabFact is a binary classification task (with labels: \textcolor{darkgreen}{Entail}, \textcolor{red}{Refute}) on Wikipedia derived tables. We use the standard train and test splits in our experiments, and report the official accuracy metric. TabFact gives simple and complex tags to each example in its test set, referring to statements derived from single and multiple rows respectively. Complex statements encompass a range of aggregation functions applied over multiple rows of table data. We report and analyze our results on simple and complex test data separately.

\subsection{Results and Analysis}
\label{sec:resultsSection}
We describe the results of our experiments with respect to the research questions outlined above.

\paragraph{1. As Evaluation benchmarks (RQ1)}

We randomly sample small subsets from each dataset, including counterfactual tables, to create test sets. We evaluate the publicly available TAPAS-TNLI model \citep{julian} fine-tuned on TabFact on the randomly sampled test sets, as shown in \autoref{tab:evaluationresults}. We find that even though TabFact contains both simple and complex training data, the model gives a best accuracy of 68.6\%, more than 12 points behind its accuracy on the TabFact set. 

\textbf{Analysis.} The TAPAS-TNLI model performs best on WikiSQL-TNLI data, showing either that WikiSQL is most comparable to TabFact (in terms of domain, reasoning, and writing) or that WikiSQL is relatively trivial to address. Squall-TNLI is the hardest, as expected, as Squall was designed specifically to include questions that execute complex SQL logic. QA-NLI and ToTTo-NLI lie in-between, showing that they have some similarities with TabFact, but also incorporate complementary reasoning instances.


\noindent \paragraph{2. Zero Shot Inference Performance (RQ2)}

Once we pre-trained our model on recasted TNLI data, it is in principle already a table NLI model. Since we create a versatile and large scale dataset, we look at the zero-shot accuracy of our models on the TabFact test set before fine-tuning, as shown in \autoref{tab:zeroshotresults}. Our best model gives 83.5$\%$ accuracy on the simple test set before fine-tuning. Its performance is 6.0$\%$ percent ahead of Table-BERT, a \textit{supervised} baseline. Our best model also outperforms TAPAS-Row-Col-Rank \cite{dong-smith-2021-structural}, which is a model trained on synthetic NLI data, by 7\% in the zero-shot setting.

\textbf{Analysis.} QA-TNLI achieves the best zero-shot performance of 71.1\%. We speculate that joining two datasets (FeTaQA and WikiTableQuestions) helps the model learn a variety of linguistic structures and reasoning. This is closely followed by Combined-TNLI, a model trained on the mixture of all the datasets. We speculate that the model's training may have been negatively impacted by integrating too many distinct data kinds. Squall-NLI noticeably gives 62.6\% accuracy on the complex test set, indicating its utility for learning complex reasoning. The zero-shot accuracy of TabFact trained models in Squall-NLI (i.e. Table \autoref{tab:evaluationresults}) and that of Squall-NLI trained model on TabFact (55.1\% vs 69.1\%) clearly show that Squall-NLI is a superior dataset in terms of complexity of reasoning. ToTTo-TNLI performs fairly well on simple data (80.3\%) but is not well equipped to handle complex examples. This is due to the ``descriptive'' nature of generation data, which includes limited inferential assertions.


\noindent \paragraph{3. Data Augmentation for TabFact (RQ3)}

Since TabFact is a binary classification task with Entail and Refute labels, our recasting data can also be used for data augmentation. We pre-train the model with our recasted data, similar to \citet{julian} (refer \ref{subsection:model}), before final fine-tuning on the TabFact dataset. \autoref{tab:augmentationresults} shows the performance after data augmentation. Our best model outperforms the Table-BERT and LPA Ranking baselines \citep{2019TabFactA} by 17 points, and \citet{julian} by 1.1 points.

\begin{table} [t]
\small
\centering
\begin{tabular}{lcc}
\hline
\textbf{Test Set} & 
\multicolumn{2}{c}{\textbf{Model}} \\
&Base&Large\\
\hline
QA-TNLI& 56.1 & 58.0\\
WikiSQL-TNLI& 66.8 &68.6\\
Squall-TNLI& 53.7 & 55.1\\
ToTTo-TNLI& 64.9 & 65.6\\
\hline
\end{tabular}
\caption{\small Accuracies for base and large TAPAS-TNLI model trained on TabFact and tested on recasted datasets}
\label{tab:evaluationresults}
\end{table}

\begin{table} [t]

\centering
\small
\begin{tabular}{lrrr}
\hline
\textbf{Model} & 
\multicolumn{3}{c}{\textbf{TabFact}} \\
&simple&complex&full(s+c) \\
\hline
Table-BERT\textsubscript{sup}&79.1&58.2&65.1 \\
LPA Ranking\textsubscript{sup}&78.7&58.5&65.3 \\
Tapas-RC-Rank&76.4&57.0&63.3 \\
\hline
QA-TNLI&\bf 83.5 &\bf 64.9 &\bf 71.1\\
WikiSQL-TNLI& 79.0 & 57.7 & 64.9\\
Squall-TNLI& 82.0 & 62.6 & 69.1\\
ToTTo-TNLI& 80.3 & 59.6 & 66.7 \\
Combined-TNLI& 83.0 & 62.9 & 69.7\\ 
\hline
\end{tabular}
\caption{\small Zero-shot accuracies for models trained on recasted data and tested on TabFact simple, complex and full dev set. Table-BERT and LPA Ranking are supervised baselines taken from TabFact \cite{2019TabFactA}. \cite{dong-smith-2021-structural} gives the zero-shot accuracy of TAPAS-Row-Col-Rank on TabFact.}
\label{tab:zeroshotresults}
\end{table}

\begin{table*}[t]
\small
\centering
\begin{tabular}{lcrrrrr}
\toprule
\multicolumn{2}{c}{\textbf{Model}} &
\textbf{Dev} & 
\textbf{Test\textsubscript{full}} & \textbf{Test\textsubscript{simple}} & \textbf{Test\textsubscript{complex}} & \textbf{Test\textsubscript{small}}\\
\hline
\multicolumn{2}{l}{Table-BERT-Horizontal\citep{2019TabFactA}} & 66.1 &65.1 & 79.1 & 58.2 & 68.1\\
\multicolumn{2}{l}{LPA-Ranking \citep{2019TabFactA}} & 65.1 & 65.3 & 78.7 & 58.5 & 68.9 \\
\multicolumn{2}{l}{Logical-Fact-Checker \citep{zhong-etal-2020-logicalfactchecker}} & 71.8 & 71.7 & 85.4 & 65.1 & 74.3\\
\multicolumn{2}{l}{HeterTFV \citep{shi-etal-2020-learn}} & 72.5 & 72.3 & 85.9 & 65.7 & 74.2\\
\multicolumn{2}{l}{Structure-Aware TF \citep{zhang-etal-2020-table}} & 73.3 & 73.2 & 85.5 & 67.2 & -\\
\multicolumn{2}{l}{ProgVGAT \citep{yang-etal-2020-program}} & 74.9 & 74.4 & 88.3 & 67.6 & 76.2\\
\multicolumn{2}{l}{TableFormer \citep{Yang2022TableFormerRT}} & 82.0 & 81.6 & 93.3 & 75.9 & 84.6 \\
\multicolumn{2}{l}{TAPAS+Salience \citep{wang-etal-2021-table-based}} & 82.7 & 82.1 & 93.3 & 76.7 & 84.3 \\
\midrule
\multicolumn{2}{l}{TAPAS + CF + Syn \citep{julian}} & 81.0 & 81.0 & 92.3 & 75.6 & 83.9\\
\multicolumn{2}{l}{QA-TNLI (Question Answering)}  & 81.4 &\bf 81.8 & 92.6 &\bf 76.4 & 84.0\\
\multicolumn{2}{l}{WikiSQL-TNLI (Semantic Parsing)} & 78.3 & 78.6 & 91.2 & 72.4 & 80.9\\
\multicolumn{2}{l}{ Squall-TNLI (Semantic Parsing)} & 80.6 & 80.5 & 91.9 & 74.9 & 82.3\\
\multicolumn{2}{l}{ToTTo-TNLI (Table2Text Generation)}  &\bf 81.9 &\bf 82.1 &\bf 93.7 &\bf 76.4 &\bf 85.4\\
\multicolumn{2}{l}{Combined-TNLI}  & 81.0 & 80.5 & 92.0 & 74.8 & 83.7\\
\midrule
Human & - & - & - & - & - & 92.1\\
\bottomrule
\end{tabular}
\caption{\small Accuracies on TabFact, including the Human Performance. Table-BERT-Horizontal and LPA-Ranking (w/ discriminator) are baselines taken from TabFact \citep{2019TabFactA}. CF means CounterFactual data, TF means TansFormers, LPA means Latent Program Algorithm. ToTTo-TNLI, QA-TNLI (WikiTQ + FeTaQA), WikiSQL - TNLI and Squall - TNLI are table NLI models pre-trained on CF + Synthetic data \citep{julian} followed by respective re-casted datasets. Combined - TNLI is a model trained on all of the data, starting with CF + Synthetic data and then mixing data from recasted datasets in equal rates.}
\label{tab:augmentationresults}
\end{table*}

\textbf{Analysis.} Following the zero-shot results (\autoref{tab:zeroshotresults}), QA-TNLI performs well as expected in the fine-tuned setting. We speculate that ToTTo-TNLI outperforms QA-TNLI due to their dataset size disparity (nearly 8x more, refer \autoref{table:dataset-stats}). The fact that WikiSQL-TNLI achieved the highest accuracy with TabFact-trained models (\autoref{tab:evaluationresults}) and the lowest zero-shot accuracy (\autoref{tab:zeroshotresults}) on TabFact indicates that the data is relatively non-complex. Squall-TNLI does not improve model performance after augmention despite its remarkable zero-shot performance (\autoref{tab:evaluationresults}). We suspect that this is because the domains and types of underlying logic (a.k.a. reasoning types) are quite distinct. We also combine all datasets (in equal rates) to train a composite TNLI model. Its accuracies are not at par with our best model. There can be several reasons behind this, one being that our mixing strategy isn't optimal. We could, for example, train for one dataset at a time and then slowly go on to the next, instead of mixing all datasets at each stage in equal proportions. This can be further investigated in the future. Another possibility is that the datasets include distinct types of data, such that merging them all has a detrimental effect.
  
\section{Related Work}
\label{sec:related_work}

\paragraph{Inference on Semi-Structured Data.} 
In recent times, inference tasks such as NLI, Question Answering and Text generation have been applied to structured data sources like tables. TabFact \cite{2019TabFactA} and InfoTabs \cite{Gupta:20} explore inference as an entailment task. WikiTableQuestions \cite{pasupat:15}, WikiQAA \cite{Abbas2016WikiQAA}, FinQA \citep{chen-etal-2021-finqa} and HybridQA \cite{chen2020hybridqa} perform question answering on tables. ToTTo \cite{parikh2020totto}, \citet{yoran2021turning}, LogicNLG \cite{chen2020logical} and Logic2Text \citep{Chen2020Logic2TextHN} explore logical text generation on tables. Most of these datasets derive tables from Wikipedia.

Early work on structured data modeling classify tables into structural categories and embed tabular data into a vector space \citep{GhasemiGol2018TabVecTV, 9005681,Deng2019Table2VecNW}. Recent work like TAPAS \citep{herzig-etal-2020-tapas}, TAPAS-Row-Col-Rank \cite{dong-smith-2021-structural}, TaBERT \cite{yin-etal-2020-tabert}, TABBIE \cite{iida-etal-2021-tabbie}, Tables with SAT \cite{zhang-etal-2020-table}, TabGCN \cite{pramanick-2021-joint} and RCI \citep{glass-etal-2021-capturing} use more sophisticated methods of encoding tabular data. TAPAS \cite{herzig-etal-2020-tapas}  encodes row/column index and order via specialized embeddings and pre-trains a MASK-LM model on co-occurring Wikipedia text and tables. \citet{yang-zhu-2021-exploring-decomposition} decomposes NLI statements into subproblems to enhance inference on TabFact.

\paragraph{Data Augmentation.}
Generating cheap and scalable data for the purpose of training and evaluation has given rise to the use of augmentation techniques. Synthetic data generation for augmentation for unstructured text is explored in \citet{alberti-etal-2019-synthetic, lewis-etal-2019-unsupervised, wu-etal-2016-bilingually, Leonandya2019TheFA}, and for Tabular NLI is shown in \citet{geva, julian}. \citet{Salvatore2019UsingSA} and \citet{dong-smith-2021-structural} generate synthetic data for evaluation purposes. Closer to our work, \citet{sellam-etal-2020-bleurt} use perturbations of Wikipedia sentences for intermediate pre-training for BLEURT(a metric for text generation) and \citet{Xiong2020PretrainedEW} replace entities in Wikipedia by others with the same type for a MASK-LM model objective. 
\paragraph{Data Recasting.} 
Data generation through recasting has been previously explored for NLI on unstructured data. \citet{white-etal-2017-inference} uses semantic classification data as their source. Multee \citep{Trivedi2019RepurposingEF} and SciTail \cite{scitail} recast Question Answering data for entailment tasks. \citet{Demszky2018TransformingQA} proposes a framework to recast QA data for NLI for unstructured text. \citet{poliak:2018} presents a collection of recasted datasets originating from seven distinct tasks. For tabular text, \citet{dong-smith-2021-structural} present an effort to re-use text generation data for evaluation. 

\paragraph{Table Pre-training.} Existing works explore pre-training through several tasks such as Mask Column Prediction in TaBERT \cite{yin-etal-2020-tabert}, Multi-choice Cloze at the Cell Level in TUTA \cite{Wang2021TUTATT}, Structure Grounding \cite{deng-etal-2021-structure} and SQL execution \cite{liu2021tapex}. Our work is closely related to \citet{julian}, which uses two pre-training tasks over Synthetic and Counterfactual data to drastically improve accuracies on downstream tasks.
Pre-training data is either synthesized using templates \citep{julian}, mined from co-occuring tables and NL sentence contexts \cite{yin-etal-2020-tabert, herzig-etal-2020-tapas}, or directly taken from human-annotated table-NLI datasets \cite{deng-etal-2021-structure, Yu2021GraPPaGP}. In our study, we employ pre-training data that has been automatically scaled from existing non-NLI data. 

\section{Conclusion}
\label{sec:conclusion}

In this paper we introduced a semi-automatic framework for recasting tabular data. We made our case for choosing the recasting route due to its cost effectiveness, scalability and ability to retain human-alike diversity in the resultant data. Finally, we leveraged our framework to generate NLI data for five existing tabular datasets. In addition, we demonstrated that our recasted datasets could be utilized as evaluation benchmarks as well as for data augmentation to enhance performance on the Tabular NLI task on TabFact \citep{2019TabFactA}.

\section{Limitations}
\label{sec:limitations}

While our work on tabular data recasting produces intriguing outcomes, we observe the following limitations in our approach:

\begin{enumerate}
\itemsep0em 
    \item Source datasets are designed for tasks different than the target. While our methodology assures that recasted data retains the strengths and positive qualities of its original source, we have observed that some of these traits may not necessarily coincide with the targeted task. For instance, generation tasks provide ``descriptions'', therefore the annotated data is \textit{descriptive} in nature, but it is unlikely to contain complicated reasoning involving common sense and table-specific knowledge. In addition, any faults in the original data (e.g. bias issue) may get transferred to the recasted version.
    
    \item Although the domains of source and target tasks can be comparable (in our example, open-domain Wikipedia tables), their distributions of categories, themes, and so on are likely to vary. When we train models using recasted augmentation data, we unintentionally introduce a domain transfer challenge. As a result, the final model's performance is influenced to some extent by domain alignment.
    
    \item Tables are semi-structured data representations that differ not just in domains and writing style, but also in structure. For example,  {InfoTabS} \citep{neeraja-etal-2021-infotabskg} is a collection of Infoboxes, which are tables that describe a single entity (person, organisation, location). These are very different from the database-style tables that we use in our research. Tables can also be chronological, nested, or segmented which makes them more challenging. While we can employ our current heuristics to identify such tables, our current recasting strategy is prone to failure with tables that do not have database-like structures.
    
    \item Annotated data sometimes relies on common sense and implicit knowledge that is not explicitly mentioned in the premise. Such data instances might be difficult to interpret automatically, making them challenging to recast. For example, in \autoref{table:limit}, to compare "Gold" with "Silver", the association of "Silver medal" with 2$_{nd}$ place and "Gold medal" with 1$_{st}$ place must be known. This implicit common-sense like knowledge makes this example hard to recast.
    
    \begin{center}
\begin{table} [h]
\small
\centering
\begin{tabular} {X|l|l|l|}

\multicolumn{3}{c}{\textbf{Micheal Phelps - 100m Butterfly}} \\ 
\hline
Vanue&Year&Medal \\
\hline
Olympics, Beijing & \cellcolor{blue!25}2008 & \cellcolor{blue!25}Gold  \\
Olympics, London & 2012 & Gold  \\
Olympics, Rio & \cellcolor{blue!25}2016  & \cellcolor{blue!25}Silver \\
\hline
\multicolumn{3}{c}{Label: \textcolor{darkgreen}{\textit{Entailment:}}} \\
\multicolumn{3}{c}{H: Micheal Phelps ranked better in 2012} \\ 
\multicolumn{3}{c}{than in 2016 for the 100m Butterfly event.} \\

\end{tabular}
\vspace{-1.0em}
\caption{\small An example table and an entailment derived from the same.}
\label{table:limit}
\end{table}
\end{center}

    \item Our work on data recasting is done only on English language data. However, our proposed framework is easily extensible to other languages, high resource and low resource alike. Since we depend on identifying and aligning entities (between premise and hypothesis), morphologically analytic languages are easier to work with. Highly agglutinative languages may require additional efforts such as morph-analysis.
\end{enumerate}

\section*{Acknowledgement}
We thank members of the Utah NLP group for their valuable insights and suggestions at various stages of the project; and reviewers their helpful comments. We thanks Chaitanya Agarwal for valuable feedback. Additionally, we appreciate the inputs provided by Vivek Srikumar and Ellen Riloff. Vivek Gupta acknowledges support from Bloomberg’s Data Science Ph.D. Fellowship.
\bibliography{anthology,custom}
\bibliographystyle{acl_natbib}

\end{document}